\documentclass[biblatex,nocompat]{trbunofficial}

\usepackage{graphicx}

\usepackage{amsmath,amsthm,amssymb}
\usepackage{optidef}
\usepackage{multirow}
\usepackage{txfonts}
\usepackage{booktabs}
\usepackage[utf8]{inputenc}

\AuthorHeaders{Rui, Mansur, and Ding}
\title{An "Xcity" Optimization Approach to Designing Proving Grounds for Connected and Autonomous Vehicles}

\author{%
  \textbf{Rui Chen}\\
  University of Michigan, Ann Arbor.\\
  500 S. State Street, Ann Arbor, MI 48109 USA\\
  Email: richen@umich.edu\\
  \hfill\break
  \textbf{Mansur Arief}\\
  Carnegie Mellon University\\
  5000 Forbes Avenue, Pittsburgh, PA 15213\\
  Email: marief@andrew.cmu.edu\\
  \hfill\break%
  \textbf{Ding Zhao, Ph.D.}\\
  Carnegie Mellon University\\
  5000 Forbes Avenue, Pittsburgh, PA 15213\\
  Email: dingzhao@cmu.edu
}

\begin{document}
\maketitle

\section{Abstract}

Proving ground, or on-track testing has been an essential part of testing and validation process for connected and autonomous vehicles (CAV). Several world-class CAV proving grounds, such as Mcity at the University of Michigan and The Castle of Waymo, have already been built, and many more are currently under construction. In this paper, we propose the first optimization approach to CAV proving ground designing and refer to any such CAV-centric design problem as ``Xcity'' to emphasize the enormous investment, the multi-dimensional spatial consideration, and the immense construction effort emerging globally. Inspired by the recent progress on traffic encounter clustering, we further define ``road assets'' as fundamental building blocks and formulate the whole design process into nonlinear optimization problems. We have shown that such framework can be utilized to adaptively generate CAV proving ground designs with optimized capability and flexibility and can further be extended to evaluate an existing ``Xcity'' design.

\vspace{1\baselineskip}
\noindent\textit{Keywords}: connected and autonomous vehicle, testing and validation, proving ground design, traffic encounter, optimization
\newpage

\section{Introduction}
\label{sec:Intro}
\noindent Development of self-driving technologies has drawn significant public attention with major concerns on safety issues. Extensive efforts have been made on testing and validation (T\&V) of both autonomy and safety of connected and autonomous vehicle (CAV) systems before deployment. The "V-model" for CAV systems \cite{IgorP2015Vmodel} illustrates the T\&V process from component level to traffic level. An approach for such process, visualized as a T\&V pyramid \cite{zszalay2016pyramid}, has identified proving ground testing as an essential level required by CAV systems. CAV proving ground is a reserved area designed and constructed to simulate real-world traffic environment. The fully controllable environment of proving grounds provides CAV testing and validation with foreseeable risks and flexible fidelities. Those advantages have led to the establishment of several world-class CAV proving grounds: Mcity at the University of Michigan \cite{mcity}, "The Castle" of Waymo \cite{castle}, Uber's ALMONO \cite{almono}, Smart Mobility Advanced Research and Test Center (SMART) by Transportation Research Center (TRC) \cite{trcsmart}, Security Smart Mobility Analysis and Research Test (SMART) Range in Israel by HARMAN \cite{harman}, American Center for Mobility (ACM)'s test facility at Willow Run in Michigan \cite{acm}, etc. Since massive time and resources have been invested in CAV proving ground construction, it is natural and essential to aim at effective and efficient designs. However, design theory of CAV proving grounds is rarely seen in the existing literature.

When designing an efficient and effective CAV proving ground, one should make clear the testing capability required by CAV T\&V starting from elementary system integration, such as path following, till high-fidelity traffic simulation, such as CAV tests. Due to the infeasiblity of complete testing on CAV systems \cite{philipK2016challenge}, it is natural for one to break multi-vehicle-multi-event and long-term CAV tests into a variety of small and distinguished use cases which represent typical usage scenarios for autonomous driving \cite{Wachenfeld2016}. Such typical use cases or scenarios are similarly identified in \cite{PATH2014, NHTSA} as behavioral competency, the ability of a CAV to operate in traffic conditions which it will regularly encounter; these conditions include keeping the vehicle in the lane, obeying traffic laws, following reasonable etiquette, and responding to other vehicles, road users, or commonly encountered hazards. Due to the highly complicated and stochastic nature of real traffic, it would be impractical to require tests for all possible use cases. Therefore, it is reasonable for one to identify minimum behavioral competencies as required. Such efforts lead to CAV capability checklists which will be practical for one to use as evaluation criterion for CAV systems \cite{PATH2014, TORC}. Thus, a CAV proving ground is considered both practical and effective when it provides wide coverage of identified behavioral competency tests within constrained space. 
To the best of our knowledge, the literature provides no such systematic approach that employs optimization models to maximize the testing capability of a CAV proving design. 
This work contributes to the formulation of a systematic proving ground design that cognizantly addresses the CAV evaluation challenges to assist the design of CAV proving grounds in terms of maximizing testing capability and flexibility.

Challenges of systematically designing an effective and efficient CAV proving ground come from two aspects: a) it is not clear how to classify and extract typical driving scenarios from a large scale naturalistic driving data; b) it is not clear how to map those scenarios into space-constrained proving grounds. To address these problems, we propose the development of Xcity: a mathematical approach to design CAV proving ground road layout based on traffic encounter theory \cite{trafficprimitive, wenshuo, sisiLi} and nonlinear optimization methods. We applied and illustrated our framework with a minimal example. Then we extracted driving scenarios from Mcity road map and from clustered driving scenarios by \cite{wenshuo}, thereby exploring the scalability of our approach.


\section{Xcity Design Framework}
\label{sec:framework}
The underlying idea of Xcity design framework is to generate a CAV proving ground road map which provides maximum coverage of typical CAV driving scenarios within constrained space and maximum flexibility of scenario transitions. In this section, we describe the systematic approach in terms of scenario extraction and representation, scenario selection, and placement optimization.

As mentioned in introduction section, we evaluate the capability of a CAV proving ground based on the CAV driving scenarios it supports. It is worth mentioning that this paper focuses on scenarios involving at least two vehicles. 
The reason behind this is that single-vehicle tests such as lane following, traffic rule obeying, and intersection navigations can be triggered by simply removing all the other road users from the corresponding multi-vehicle scenarios, e.g. car-following including Stop \& Go \& Emergency Stop and 2-way stops obeying with cross traffic \cite{TORC, PATH2014}. Thus, we argue that the evaluation requirements of single-vehicle tests can be trivially incorporated whenever more complicated multi-vehicle scenarios are conductable.

Therefore, it is essential to first develop a systematic way to identify, extract, and represent typical multi-vehicle CAV usage cases from a large-scale driving data. Such task is non-trivial since the real-world traffic is a vast, stochastic, and dynamic cyber-physical system consisting of large number of road users, non-road users, and stochastic environment factors \cite{trafficprimitive}. The method developed in \cite{sisiLi, wenshuo, trafficprimitive} allows the processing of massive multidimensional traffic data to extract traffic primitives and traffic encounters. In this context, traffic primitives, which are the most informative traffic features from the dataset, are then used as principal compositions of the entire traffic \cite{trafficprimitive}. Traffic encounters, which represent the scenario where two or multiple vehicles are spatially close to and interact with each other on roadways \cite{wenshuo}, allow higher fidelity representation of sophisticated driving scenarios. 

Our design approach aims to map such scenarios into the given space in a way that maximizes CAV evaluation capability. 
To enable tests on a particular scenario, we map the group of roads and intersections which supports the realization of traffic encounters into the proving ground design. Additional factors such as other road users, traffic control devices, and control commands are then embedded on demand according to the test specifications. For example, one could add two stops signs and deploy one other vehicle at an X-intersection to support the scenario of 2-way stop-obeying with cross traffic. With the same physical layout, one could also evaluate any scenarios of the form 
left-turn-on-green with incoming traffic. Therefore, we consider the combination of roads and intersections 
as basic construction element in our design formulation and refer to such combination as \textit{road assets} hereafter. Once selected and mapped into the design, a \textit{road asset} $a$ will be treated as a 2D rigid body afterwards and all corresponding testing scenarios it supports (based on traffic-primitives analysis) will be enabled. Furthermore, we introduce a value metric $\upsilon:a\rightarrow\mathbb{R}$ to evaluate \textit{road assets} since the number of supported scenarios for different assets could vary. For a \textit{road asset} $a$, its value comes from both its versatility to support multiple scenarios and the universality of the supported scenarios.

Given a preprocessed set of road assets $A = \{a_1, a_2,..., a_{NA}\}$, value measure $\upsilon$, and constrained construction space $S$, we propose the following constraints, definitions, and objectives to model an Xcity design problem.\\

\noindent\textit{Constraints}:
\begin{enumerate}
  \item Roads in any single asset $a_i$ cannot overlap with roads in other assets.
  \item Road assets should completely reside within the constrained space $S$.
\end{enumerate}

\noindent\textit{Definitions}:
\begin{enumerate}
  \item A subset, $A^\prime = \{a^\prime_1, a^\prime_2,..., a^\prime_{NA^\prime}\}$, of $A$ is considered feasible if all road assets in $A^\prime$ can be mapped into $S$ simultaneously under the above constraints.
  \item Two road assets $a_i$ and $a_j$ are considered directly connectible if it is possible to construct a straight transition road to connect them without colliding with any other assets. The number of directly connectible asset pairs $(a_i, a_j)$ in a road asset set $A$ is referred to as \textit{direct connectivity} hereafter and denoted as $\mathit{C}:A\rightarrow\mathbb{R}$.
\end{enumerate}

\noindent\textit{Objectives}:
\begin{enumerate}
  \item The feasible subset $A^*$ with highest total asset value $\upsilon(A^*)$ is selected and mapped in $S$.
  \item The assets in $A^*$ are arranged within $S$ such that $\mathit{C}(A^*)$ is maximized.\\
\end{enumerate}

\textit{Constraint 1} guarantees that no test fidelity is lost due to the physical design. Each scenario test should be conducted in its originally dedicated road asset with unaltered roads and intersections. Merging distinguished assets would generate undesired and inaccurate physical environment, which attenuates the fidelity of the simulated scenario. It would consequently cause non-measurable degeneration to the confidence of public street deployment gained from on-track tests.

As previously mentioned, one criteria which we use to evaluate a CAV proving ground is its coverage of behavioral competency (BC) tests. BC is defined as a list of independent scenarios that are expected to be handled safely by a sophisticated autonomous driving system. Since BC implies no connections between different scenarios, we treat the mapped \textit{road assets} independently and simply accumulate their value measures to form \textit{Objective 1}. On the other hand, Proving ground test and validation, a comprehensive and multi-integration-level process, calls for not only a wide range of independent scenario-based tests but also cascaded and potentially randomly ordered scenarios to assess CAV systems' robustness and performance on enduring operations. Thus, the success in emulating scenario sequences further boosts the flexibility and fidelity of the proving ground. Tests involving multiple independent but connected scenarios highly demand that the CAV proving ground be flexible on navigating among different scenarios. In our case, such flexibility will be ensured by transition roads built to connect different road assets. Although roads with arbitrary shapes and lengths would highly likely to serve as transitions between any pair of road assets, it is worth looking for a design with more directly connectible asset pairs. 
Therefore, we only consider straight transitions for evaluation purpose. The flexibility and fidelity measure depending on possible straight transition roads are defined as \textit{Objective 2}. Note that the the outcome of this paper will be a high-fidelity graph representation of the road map structure. Detailed information on road construction, such as exact road curvatures, lane widths, and sidewalks, needs to be added to our design, so as to achieve a final construction blueprint. Further discussion on blueprint rendering is out of scope and is omitted. 

\section{Methodology}

In this section, we describe the mathematical formulation and optimization models of our approach. With \textit{road assets} pre-produced and represented, the design process is decomposed into two phases. In phase 1, the feasible \textit{road asset} subset $A^*$ with highest total asset value $\upsilon(A^*)$ is selected from all pre-produced \textit{road assets} $A$. In phase 2, the placement of the pre-selected $A^*$ within given space $S$ will be optimized in terms of cross-scenario transition flexibility. Both two phases are modeled into nonlinear optimization problems.

\subsection{\textit{Road Asset} Representation}

Traffic encounters are clustered as GPS location sequences of involved vehicles. Thus, given a traffic encounter, one natural and direct approach to retrieve the essential \textit{road asset} is to directly locate the traffic encounter on the world map and extract surrounding roads and intersections. To map the extracted \textit{road assets} into the given space, we will need to define a geometric descriptor for their structure and size. On one hand, such descriptor should be rigorous enough for accurate localization and space-reserving for the corresponding asset. On the other hand, the descriptor should be concise enough to truncate unnecessary computations during optimization. Two of the most recognized map representations for autonomous driving are Lanelets \cite{lanelets} and OpenDRIVE \cite{opendrive}. Both of them are efficient and detailed enough for semantic environment modeling and motion planning. However, our approach only aims at the placement and transitions of \textit{road assets} regardless of their semantic meanings or navigational information. These knowledge is already contained in the supported scenarios and can be applied on demand after the design is finished. Therefore, both Lanelets and OpenDRIVE contain more information than needed and create unnecessary computations.

\begin{figure}[!ht]
  \centering
  \includegraphics[width=0.9\textwidth]{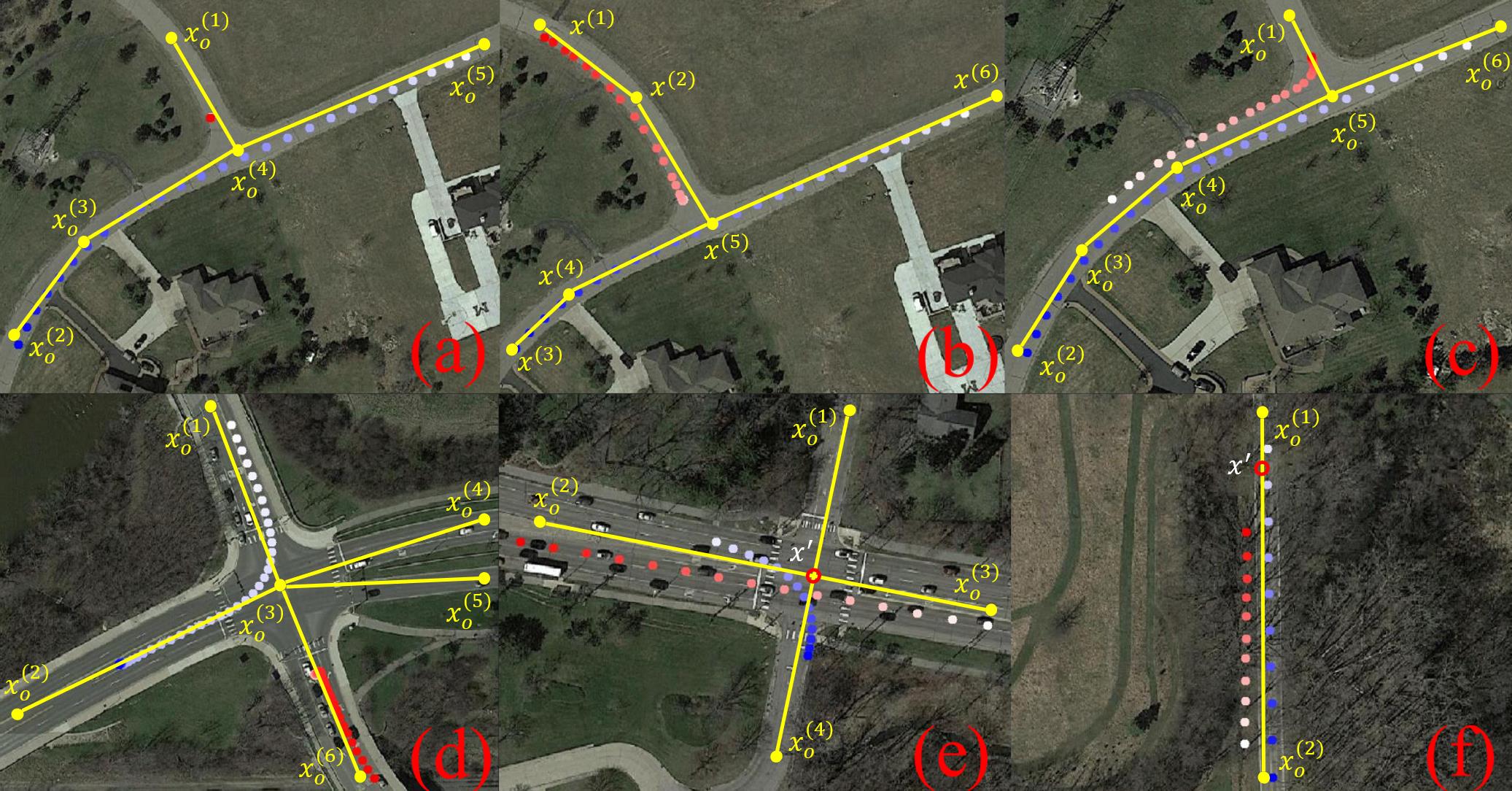}
  \caption{Examples of \textit{road asset} structure graph and corresponding extracted traffic encounters. Deep red and blue dots are the start positions of vehicles. White dots are the end positions of vehicles \cite{wenshuo}. Yellow dots refer to nodes, which are connected by yellow lines representing \textit{internal segments}. Nodes are labeled with their indexes within the assets. Red circles with white $x^\prime$ label, in (e)-(f), mark nodes that are ignored for graph simplification.}\label{fig:AssetEnc}
\end{figure}

Considering both representativeness and simplicity, we use graphs to model the geometric structure of \textit{road assets} (see examples in Figure \ref{fig:AssetEnc}). Such descriptor is inspired by the map format from OpenStreetMap (OSM) \cite{OpenStreetMap}, a collaborative project to create a free editable map of the world \cite{wiki:osm}. We mark the intersections, road ends, and way-points along curved roads with nodes $x^{(i)}$. Then, we connect each node pair $(x^{(i)},x^{(j)})$, if necessary, with a line segment $\ell(x^{(i)},x^{(j)})$ to model one road segment in the real world. We refer to those line segments as the \textit{internal segments} of \textit{road assets}. We ignore nodes that are unnecessary or trivial, thereby maintaining the whole structure to introduce the least possible \textit{internal segments} and save computation efforts. Typical trivial nodes include nodes at X-intersection centers (see Figure \ref{fig:AssetEnc}.e). Notably, the number of \textit{internal segments} turns out to be one major factor contributing to model complexity.

\subsection{Phase 1: Most-valuable Feasible Subset Selection Model}

\subsubsection{\textit{Single-Asset Constraint Set} (SACS)}

We map \textit{road assets} into the given space $S$ by mapping their corresponding graph representations while retaining the relative placement of the nodes. This is done through the introduction of a set of 2D geometric constraints on the nodes of which the original node placement is the only realization. A non-trivial start point would be mapping a triangle defined by the original vertex positions $x^{(i)}_o, x^{(j)}_o, x^{(k)}_o\in\mathbb{R}^{2}$ into space $S$ at $x^{(i)}, x^{(j)}, x^{(k)}\in\mathbb{R}^{2}$ without deformation or flipping. We propose the following constraint set $K_\Delta(x^{(i)}, x^{(j)}, x^{(k)})$, referred to as \textit{triplet constraint set}, which retains the triangle's original shape and size. Note that the superscript number refers to the node index within a single asset. The subscript 'o' refers to the original location of nodes on world map.
\begin{linenomath}
  \postdisplaypenalty=0
    \begin{align}
      & dist(x^{(i)}, x^{(j)})^2 == dist(x^{(i)}_o, x^{(j)}_o)^2 + \delta_{ij} \label{eq:d1} \\
      & dist(x^{(i)}, x^{(k)})^2 == dist(x^{(i)}_o, x^{(k)}_o)^2 + \delta_{ik} \label{eq:d2} \\
      & dist(x^{(j)}, x^{(k)})^2 == dist(x^{(j)}_o, x^{(k)}_o)^2 + \delta_{jk} \label{eq:d3} \\
      & b_{CCW} = \mathbb{I}\Big[\Omega\left(x_o^{(i)}, x_o^{(j)}, x_o^{(k)}\right) > 0\Big] \label{eq:bCCW}\\
      & (-1)^{b_{CCW}}\Omega\left(x^{(i)}, x^{(j)}, x^{(k)}\right) \leq 0 \label{eq:ornt} \\
      & x^{(i)}, x^{(j)}, x^{(k)} \in S \label{eq:inS}
    \end{align}
\end{linenomath}
\noindent $\Omega(x, y, z)$, the \textit{orientation test}, is positive if $x,y,z\in\mathbb{R}^{2\times1}$ are placed counter-clock-wise (CCW), or negative if clock-wise (CW), or zero if collinear (LNR) \cite{orient}. See Equation \ref{eq:orientdef} for its definition. The $\delta$ variables in constraint \ref{eq:d1}-\ref{eq:d3} are introduced as slack variables for equality constraints to avoid numerical issues in solvers. The sum of all squared slack variables will be include in the objective to be minimized.
\begin{linenomath}
\postdisplaypenalty=0
  \begin{align}
    &\Omega(x,y,z)=det\left(\left[ \begin{array}{ccc}
    x^T & 1 \\
    y^T & 1 \\
    z^T & 1 \end{array}
    \right]\right) \label{eq:orientdef}
  \end{align}
\end{linenomath}

\noindent The relative placement of all nodes $x_\tau^{(1)}, x_\tau^{(2)},..., x_\tau^{(n_\tau)}\in\mathbb{R}^{2}$ in road asset $a_\tau$ can be retained by applying $K_\Delta(x^{(1)}, x^{(2)}, x^{(k)}), k\geq 3$ repeatedly to nodes in $a_\tau$. We refer to the constraint set which regulates the relative placements of all nodes within a single asset $a_\tau$ as \textit{Single-Asset Constraint Set} (SACS), denoted by $K_{sa}(a_\tau)$ (see constraint set \ref{cons:sa}).
\begin{linenomath}
\postdisplaypenalty=0
  \begin{align}
    & K_{sa}(a_\tau) := \Big\{K_\Delta(x^{(1)}, x^{(2)}, x^{(k)})\  \mid\  k = 3,4,...,n_\tau \Big\} \label{cons:sa}
  \end{align}
\end{linenomath}

\begin{figure}[!ht]
  \centering
  \includegraphics[width=1\textwidth]{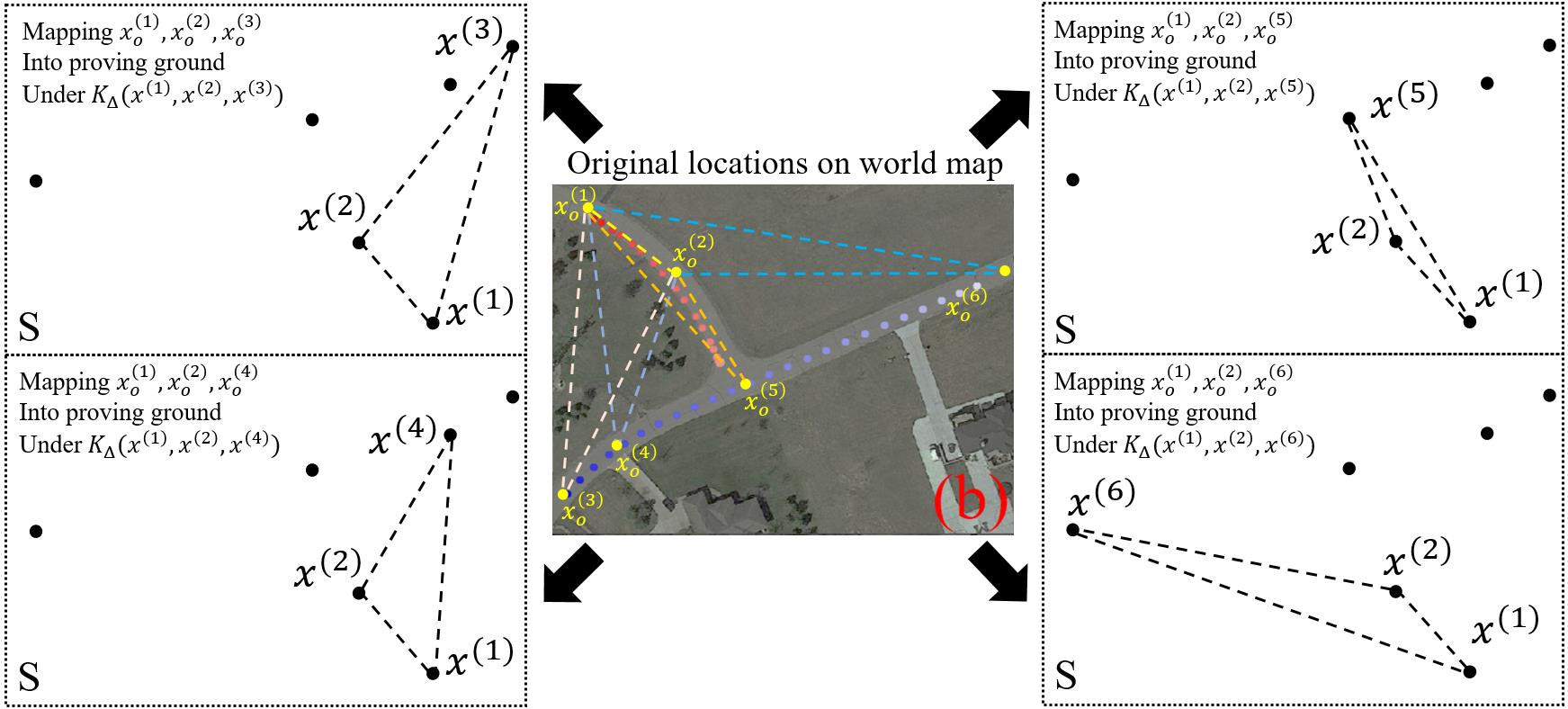}
  \caption{Example of mapping a single \textit{road asset} into space $S$. With $x^{(1)}$ and $x^{(2)}$ satisfying their distance constraint, each of the remaining nodes $x^{(k)}$ can be located uniquely by applying a \textit{triplet constraint set} with respect to $x^{(1)}$, $x^{(2)}$, and $x^{(k)}$.}\label{fig:Ksa}
\end{figure}

\subsubsection{\textit{Cross-Asset Constraint Set} (CACS)}

\begin{figure}[!ht]
  \centering
  \includegraphics[width=0.65\textwidth]{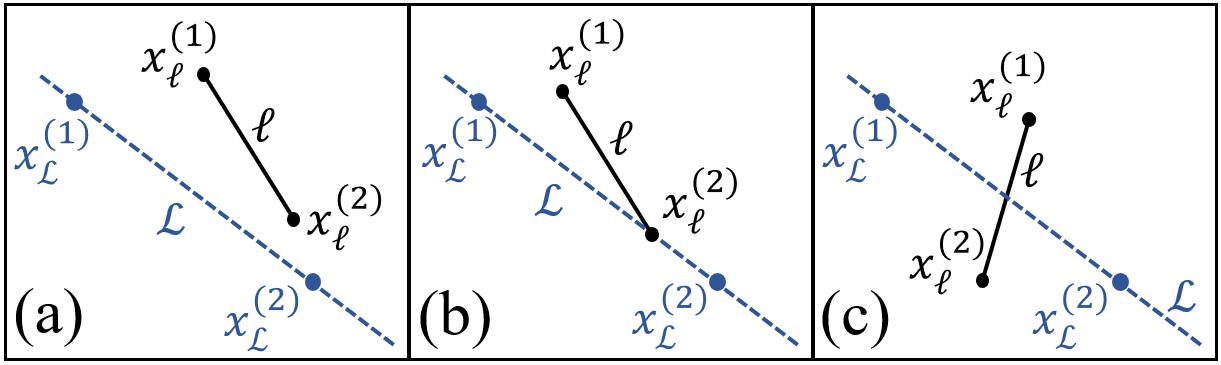}
  \caption{Examples of \textit{intersection test} with straight line $\mathcal{L}$ and line segment $\ell$. (a) $\Omega\left(x_\mathcal{L}^{(1)}, x_\mathcal{L}^{(2)}, x_\ell^{(1)}\right) > 0,\ \Omega\left(x_\mathcal{L}^{(1)}, x_\mathcal{L}^{(2)}, x_\ell^{(2)}\right) > 0 \Rightarrow \chiup(\mathcal{L}, \ell) > 0$. (b) $\Omega\left(x_\mathcal{L}^{(1)}, x_\mathcal{L}^{(2)}, x_\ell^{(2)}\right) = 0 \Rightarrow \chiup(\mathcal{L}, \ell) = 0$. (c) $\Omega\left(x_\mathcal{L}^{(1)}, x_\mathcal{L}^{(2)}, x_\ell^{(1)}\right) > 0,\ \Omega\left(x_\mathcal{L}^{(1)}, x_\mathcal{L}^{(2)}, x_\ell^{(2)}\right) < 0 \Rightarrow \chiup(\mathcal{L}, \ell) < 0$. }\label{fig:IntTest}
\end{figure}

Another requirement of road asset mapping is that \textit{internal segments} of different road assets should not collide or intersect. To formulate such requirement into constraints, we first define the \textit{intersection test} $\chiup(\mathcal{L}, \ell)$ for straight line $\mathcal{L}$ passing points $x_\mathcal{L}^{(1)}, x_\mathcal{L}^{(2)}\in\mathbb{R}^2$ and line segment $\ell$ formed by points $x_\ell^{(1)}, x_\ell^{(2)}\in\mathbb{R}^2$ as Equation \ref{eq:intTest}.
\begin{linenomath}
\postdisplaypenalty=0
  \begin{align}
    & \chiup(\mathcal{L}, \ell) = \Omega\left(x_\mathcal{L}^{(1)}, x_\mathcal{L}^{(2)}, x_\ell^{(1)}\right)\Omega\left(x_\mathcal{L}^{(1)}, x_\mathcal{L}^{(2)}, x_\ell^{(2)}\right) \label{eq:intTest}
  \end{align}
\end{linenomath}
\noindent $\chiup(\mathcal{L}, \ell)$ is then positive if $\ell$ doesn't intersect with $\mathcal{L}$ and non-positive otherwise. See Figure \ref{fig:IntTest} for illustrative examples.

Given two \textit{internal segments} $\ell_p\equiv\ell(x_p^{(1)}, x_p^{(2)})\in a_{i}$ and $\ell_q\equiv\ell(x_q^{(1)}, x_q^{(2)})\in a_{j}$, we denote the straight lines passing $(x_p^{(1)}, x_p^{(2)})$ and $(x_q^{(1)}, x_q^{(2)})$ as $\mathcal{L}_p\equiv\mathcal{L}(x_p^{(1)}, x_p^{(2)})$ and $\mathcal{L}_q\equiv\mathcal{L}(x_q^{(1)}, x_q^{(2)})$ respectively. Then $\ell_1$ doesn't intersect with $\ell_2$ ($\ell_1 \cap \ell_2 = \varnothing$) if Constraint \ref{cons:segintsct} holds.
\begin{linenomath}
\postdisplaypenalty=0
  \begin{align}
    \chiup(\mathcal{L}_p, \ell_q) > 0\ \lor\ \chiup(\mathcal{L}_q, \ell_p) > 0. \label{cons:segintsct}
  \end{align}
\end{linenomath}
\noindent To simplify the constraints, we skip considering the corner case where $\ell_p$ and $\ell_q$ are collinear but not overlapped. Such case can be trivially made feasible under our constraints by rotating or moving one of the line segments by a tiny amount. It is worth mentioning that, since strict inequalities are not permitted in most solvers, we implement constraint \ref{cons:segintsct} using non-strict equalities with a constant offset $\epsilon = 10^{-4}$. Auxiliary binary variables $b_1, b_2 \in \{0,1\}$ are introduced as well to decompose the logic OR statement into a set of AND statements for each instance of constraint \ref{cons:segintsct}. Constraint \ref{cons:segintsct} is finally formed as constraint set \ref{cons:ca}. We refer to such constraint set that eliminates the collision of $\ell_p\in a_{i}$ and $\ell_q\in a_{j}$ as \textit{Cross-Asset Constraint Set} (CACS), denoted as $K_{ca}(\ell_p,\ell_q)$.
\begin{linenomath}
  \postdisplaypenalty=0
    \begin{align}
      & K_{ca}(\ell_p,\ell_q):\quad b_1\cdot\chiup(\mathcal{L}_p, \ell_q) \geq b_1\cdot\epsilon,\quad b_2\cdot\chiup(\mathcal{L}_q, \ell_p) \geq b_2\cdot\epsilon,\quad b_1 + b_2  \geq 1 \label{cons:ca}
    \end{align}
\end{linenomath}

\subsubsection{Phase 1 Optimization Model}

With both SACS and CACS defined, we can formulate phase 1 in a single optimization model (see Expression \ref{model:phase1}). For each subset $A^\prime = \{a^\prime_1, a^\prime_2,..., a^\prime_{NA^\prime}\}\subseteq A$, let $X = \Big\{x_1^{\prime(1)},\ x_1^{\prime(2)},\ ...\ ,\ x_1^{\prime(n_1)},\\ \ x_2^{\prime(1)},\ x_2^{\prime(2)},\ ...\ ,\ x_2^{\prime(n_2)},\ ...\ ,\ x_{NA'}^{\prime(1)},\ x_{NA'}^{\prime(2)},\ ...\ ,\ x_{NA'}^{\prime(n_{NA'})}  \Big\}$ be the collection of all mapped nodes in space $S$ from $A'$. Let $\delta\in\mathbb{R}^{N_\delta\times 1}$ be a column vector containing all slack variables introduced in distance equality constraints. Let $b_{CACS}$ be the set of binary variables introduced in CACSs.
\begin{linenomath}
  \postdisplaypenalty=0
    \begin{mini!}|l|[2]
      {A'\subseteq A,X,b_{CACS}}{\delta^{\mathit{T}}\delta - \upsilon(A')}{\label{model:phase1}}{}
      \addConstraint{K_{sa}(a_\tau)}{\label{model:phase1-1}}{\quad\tau=1,2,\ldots,NA'}
      \addConstraint{K_{ca}(\ell_p,\ell_q)}{\label{model:phase1-2}}{\quad\forall\ \ell_p\in a'_{i},\ \ell_q\in a'_{j}\Rightarrow a'_i,\ a'_j\in A^\prime,\ i\neq j}
   \end{mini!}
\end{linenomath}
Notably, we factor out the selection of asset subset $A'$ during implementation and manually assign the \textit{road assets} to be mapped. When global minimum of the remaining objective $\delta^\mathit{T}\delta$ is reached, all distance Constraints \ref{eq:d1}-\ref{eq:d3} are strictly satisfied. The corresponding $X$ will be a feasible mapping of $A'$ into $S$.

\subsection{Phase 2: Transition Flexibility Optimization Model}

With the most valuable feasible subset $A^*$ selected in phase 1, we will find a placement $X^*$ of all containing \textit{road assets} such that the \textit{direct connectivity}, $\mathit{C}(A^*)$, is maximized.

\subsubsection{Asset Transition}
\begin{figure}[!ht]
  \centering
  \includegraphics[width=0.6\textwidth]{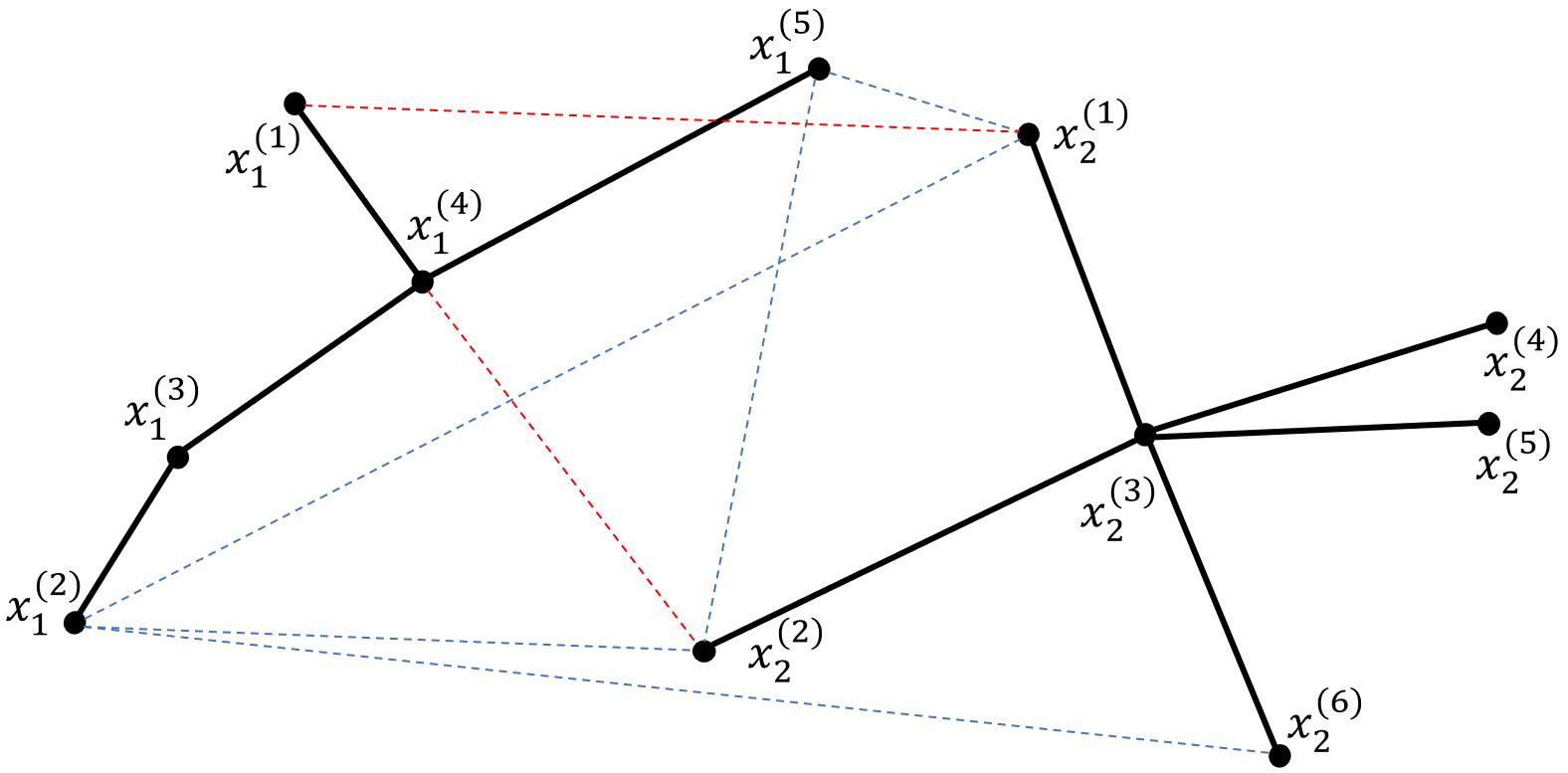}
  \caption{Example of possible transition roads between two \textit{road assets}. Blue dashed lines indicate possible transition roads. Red dashed lines indicate infeasible transition roads due to use of \textit{internal nodes} ($x_1^{(4)}$) and intersection with \textit{internal segments} ($\ell(x_1^{(4)}, x_1^{(5)})$).}\label{fig:trans}
\end{figure}

In the graph representation of a \textit{road asset}, we define nodes that are owned by only one \textit{internal segment} as \textit{boundary nodes} (for instance, $x^{(1)}$, $x^{(2)}$, $x^{(5)}$ in Figure \ref{fig:AssetEnc}.a) and the rest as \textit{internal nodes}. To avoid causing bias to the road layout and achieve high-fidelity scenarios, we only allow \textit{road assets} to be connected to each other using \textit{boundary nodes} and skip the \textit{internal nodes} when considering transition road construction. Two \textit{road assets} are considered directly connectible if it is possible to construct at least one straight transition road between them using only \textit{boundary nodes} without intersecting any \textit{internal segments} of any \textit{road asset}. See Figure \ref{fig:trans} for a mini example.

\subsubsection{Phase 2 Optimization Model}
Given a the pre-selected best feasible asset set $A^*$ from phase 1, the objective is to maximize $C(A^*)$ subject to both SACS and CACS. Given a \textit{road asset} $a_i$ with nodes $\Big\{x_i^{(1)}, x_i^{(1)},\ldots,x_i^{(n_i)}\Big\}$, we propose the following definitions. Note that the superscript * indicates that the variable is pre-selected in phase 1 and kept constant in phase 2.
\begin{itemize}
\item $b_{dc} = \{b_{dc}(a^*_i, a^*_j)\ \mid\ \forall a^*_i, a^*_j\in A^*, i\neq j\}$: set of binary decision variables indicating the direct connecting status of $a_i$ and $a_j$.
\item $\beta(a^*_i)$: set of \textit{boundary node} in $a_i$.
\item $\mathit{L}_{tr}(A^*)$: set of transition road candidates $\Big\{\ell_{tr}(x_i^{(p)}, x_j^{(q)})\ \mid\ x_i^{(p)}\in\beta(a^*_i),\ x_j^{(q)}\in\beta(a^*_j),\  i\neq j\Big\}$.
\item $b_{tr}(x_i^{(p)}, x_j^{(q)})$: binary variable which indicates the selection status of $\ell_{tr}(x_i^{(p)}, x_j^{(q)})$.
\item $\mathit{L}_{\chiup}(\ell_{tr})$: set of \textit{internal segments} that don't have common node with $\ell_{tr}$ by construction.
\end{itemize}
\noindent With SACS $K_{sa}$ and CACS $K_{ca}$ defined in Equation \ref{cons:sa} and \ref{cons:ca}, we can write the complete optimization model as Expression \ref{model:phase2}.
\begin{linenomath}
  \postdisplaypenalty=0
    \begin{mini!}|l|
      {X, b_{dc}, b_{tr}, b_{CACS}}{\quad\quad\quad\quad\quad\delta^{\mathit{T}}\delta - sum(b_{dc})\ \mid\ A^*}{\label{model:phase2}}{}
      \addConstraint{K_{sa}(a_\tau)}{\label{model:phase2-1}}{\quad\tau=1,\ldots,NA^*}
      \addConstraint{K_{ca}(\ell_p,\ell_q)}{\label{model:phase2-2}}{\quad\forall\ \ell_p\in a^*_{i},\ \ell_q\in a^*_{j}\Rightarrow a^*_i,\ a^*_j\in A^*,\ i\neq j}
      \addConstraint{\sum_{p,q}b_{tr}(x_i^{(p)}, x_j^{(q)})\geq 1}{\label{model:phase2-3}}{\quad\forall a^*_i,\ a^*_j\in A^*\Rightarrow i\neq j,\ b_{dc}(a^*_i, a^*_j)_{\mathit{true}}}
      \addConstraint{K_{ca}(\ell_{tr},\ell_\chiup)}{\label{model:phase2-4}}{\quad\forall\ell_{tr}\Rightarrow b_{tr}(\ell_{tr})_{\mathit{true}},\ \forall\ell_\chiup\in\mathit{L}_{\chiup}(\ell_{tr})}
   \end{mini!}
\end{linenomath}
\noindent Constraints \ref{model:phase2-1} and \ref{model:phase2-2} ensure that the feasibility of the mapping as identically done in phase 1. Constraint \ref{model:phase2-3} ensures that for each pair of assets to be connected, at least one transition road candidate connecting their \textit{boundary nodes} should be selected. Constraint \ref{model:phase2-4} ensures that each selected transition road candidate should not intersect with any other \textit{internal segment} unless they share a node by construction. In Figure \ref{fig:trans}, $\ell(x_1^{(4)}, x_1^{(5)})$ to $\ell_{tr}(x_1^{(5)}, x_2^{(1)})$ is one example of such exception.

\section{Optimization Approaches}
Based on our formulation, the resulting model is NP hard, essentially due to the extensive use of binaries to ensure the validity of the asset shapes and their combined designs. 
From (\ref{model:phase1}) and (\ref{model:phase2}), it is clear that even though the final objective function in (\ref{model:phase2}) is arguably convex, some of the constraints in both (\ref{model:phase1}) and (\ref{model:phase2}) are not. As such, the model is of mixed integer programming (IP) type and the natural approaches to numerically deal with it take on some form of branch-and-bound (BNB) methods. Loosely speaking, the computation procedure of BNB requires lower- and upper-bound governing how variable branches are determined to obtain integer solution at each numerical iteration \cite{lawler1966branch}. Modern solvers employ various numerical scheme deemed best to solve these branch-level problems, e.g. BMIBNB use linear relaxations to solve for the bounds from the branch problems \cite{lofberg2004yalmip}. Furthermore, other solvers have gained improved efficiencies from more sophisticated techniques to assist the branching, e.g. SCIP \cite{scip1} employs branch-cut-and-price (BCP) while BARON \cite{baron1} uses branch-and-reduce (BAR). In what follows, we describe the performance of these three solvers in regard to our formulation.

BMIBNB is a YALMIP-based BNB technique that can be easily implemented using standard computing environment Matlab \cite{lofberg2004yalmip}. YALMIP itself is an interface program that is originally designed to allow Matlab to solve semi definite programming problems using external solvers including SeDuMi and Mosek, but its implementation has currently included wider array of choices including fmincon, SDPT3, Gurobi, and CPLEX which can be tuned easily \cite{lofberg2004yalmip} to obtain the numerical bounds. In dealing with nonlinear IP, BMIBNB uses linear relaxation and convex envelope approximations. 
As such, it trades the accuracy of solution optimality for the  efficiency of computation. With this view, one shall notice that when the original problem is non-convex and high-dimensional, then the performance of BMIBNB suffers from yet-to-converge bounds. Unfortunately, our Xcity design problem is one such program and thus, as expected, the corresponding numerical experiment advises to using more customized solvers for our formulation.

SCIP 
is a BNB-based solver for integer programming enhanced with BCP framework which was designed to handle large-scale discrete optimization problem. The cuts and variable can be generated dynamically through the branching process to explore the direction of optimality. This dynamic `exploration' capability is achieved through the implementation of column generation technique, which further implies SCIP has adopted the large-scale Dantzig-Wolfe decomposition method \cite{scip1, scip2}. Furthermore, the most recent version has included Bender's decomposition, allowing the solver to efficiently deal with problems with block-angular characteristics \cite{geoffrion1972generalized}. 
Our preliminary numerical experiment shows that the SCIP's inherent column generation does not yield an appealing performance due to the insurmountable growth of the constraint size in both the number and shape-complexity of the assets.

BARON (Branch-And-Reduce Optimization Navigator) is designed to solve a generalized nonlinear programs including mixed integer nonlinear problem \cite{baron1, baron2}. The underlying idea of this technique is that after branching, a numerical analysis is performed to reduce the potential optimal regions which in turn allows a faster computation, and if conducted appropriately, global optimality assurance \cite{kilincc2018exploiting}. This notion of `appropriateness' is achieved by adapting the cutting plane and probing techniques from mixed integer linear programming to mixed integer \textit{nonlinear} problems. We refer interested readers to \cite{kilincc2018exploiting} for extensive computational experiment investigating the effectiveness of BARON scheme to achieve optimality. Our numerical results show that BARON indeed yields to much better performance both in terms of improved computational efficiency and solution quality compared to BMIBNB and SCIP. As such, we will use BARON for our further analysis in this study. It is noted that BARON requires proper license in order to be deployed.

\section{Experiments and Results}
\subsection{\textit{Road Asset} Collection}

\begin{figure}[!ht]
  \centering
  \includegraphics[width=1\textwidth]{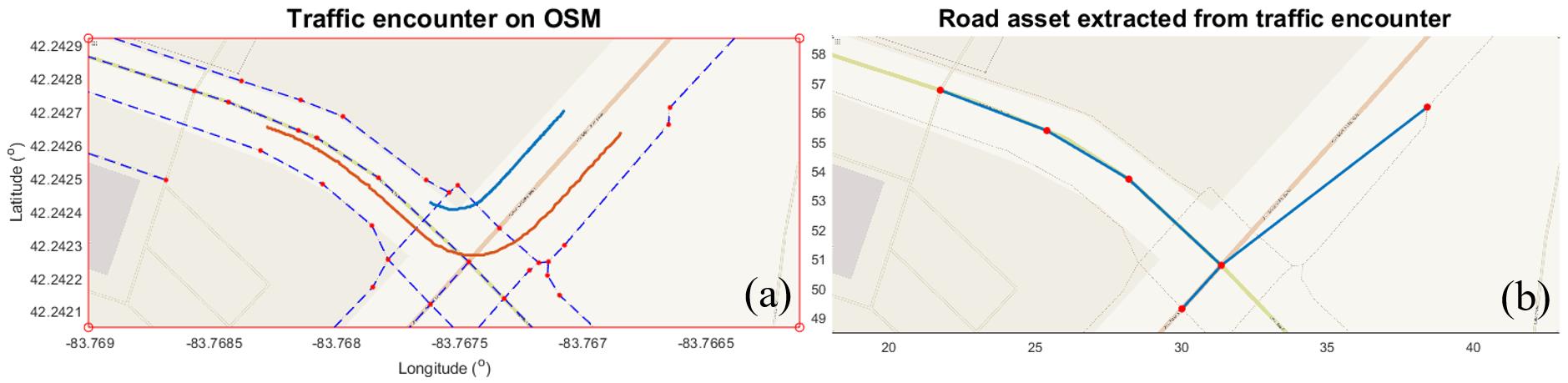}
  \caption{Example of extracted \textit{road asset} from a traffic encounter by \cite{wenshuo}. (a) Orange and blue curves refer to GPS trajectory of both vehicles. Blue dashed lines indicate OSM-logged roads and sidewalks. Red dots refer to nodes. (b) \textit{Road asset} is extracted based on available nodes. Red dots refer to nodes. Solid lines represent \textit{internal segments}}\label{fig:enc233}
\end{figure}

\begin{figure}[!ht]
  \centering
  \includegraphics[width=1\textwidth]{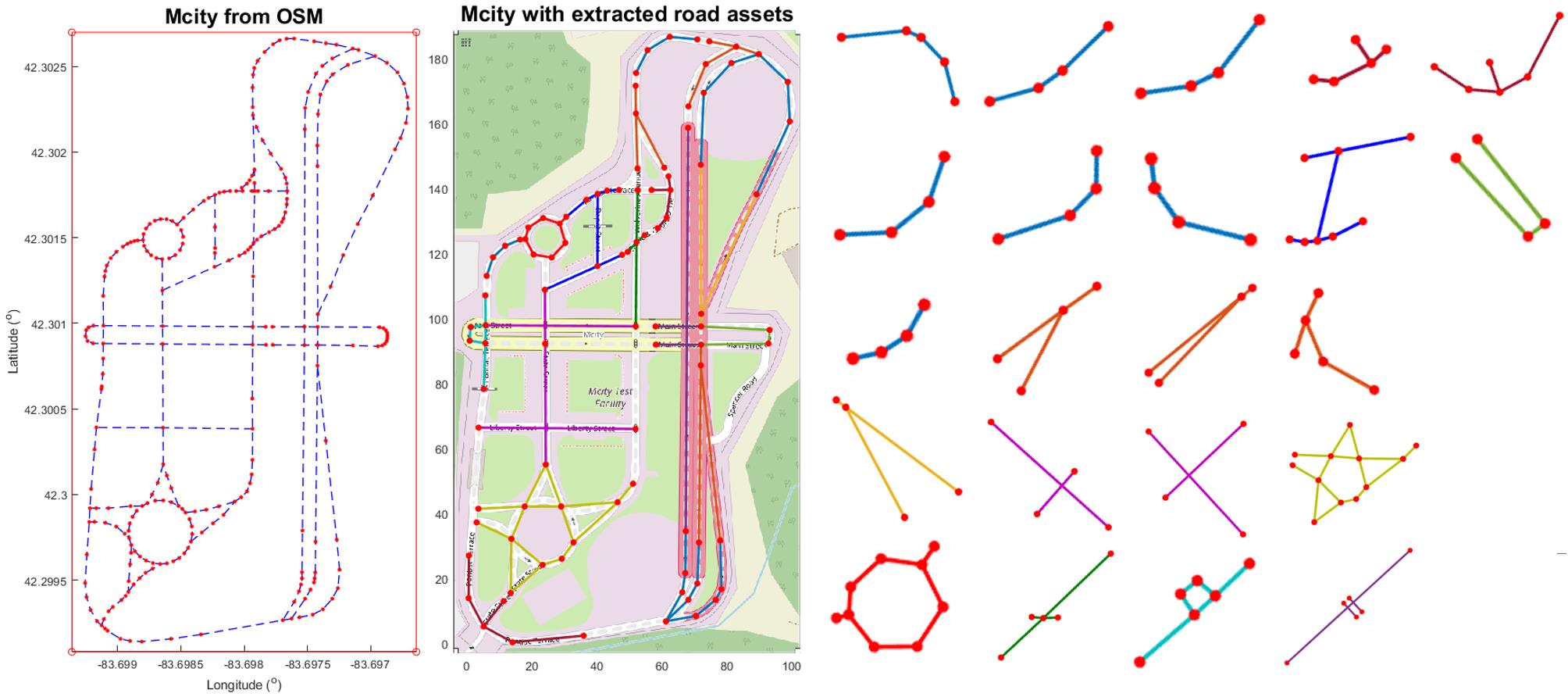}
  \caption{Left: OSM-logged Mcity road map \cite{OpenStreetMap}. Blue dashed lines indicate roads and sidewalks. Red dots refer to nodes. Middle: OSM map of Mcity overlapped with extracted \textit{road assets}. Right: All \textit{road assets} extracted. \textit{Road assets} that support common types of scenarios are shown in the same color.}\label{fig:mcity}
\end{figure}

From totally 49,998 vehicle encounters recorded by the University of Michigan Safety Pilot Model Development (SPMD) program, \cite{wenshuo} selected 2,568 naturalistic driving encounters with adequate duration. 10 typical driving scenarios were then successfully extracted. Since there is currently no method that can analytically map traffic encounters to necessary roads and intersections, we extract \textit{road assets} through manual annotation on OSM-logged road maps. After locating typical driving encounters on OSM via GPS locations, we exported the necessary surrounding road structure in .osm format and imported it into MATLAB using \cite{mathworks:osm} (see Figure \ref{fig:enc233}). Besides, we export the OSM road map of CAV proving ground at Mcity \cite{mcity} and segmented it into 22 \textit{road assets} which support different typical driving scenarios (see Figure \ref{fig:mcity}). We down-sampled the nodes in OSM raw data to simplify graph representations and assign 
random integer to value each \textit{road asset}.

\subsection{Solver Performance on Phase 1}

\begin{figure}[!ht]
  \centering
  \includegraphics[width=0.9\textwidth]{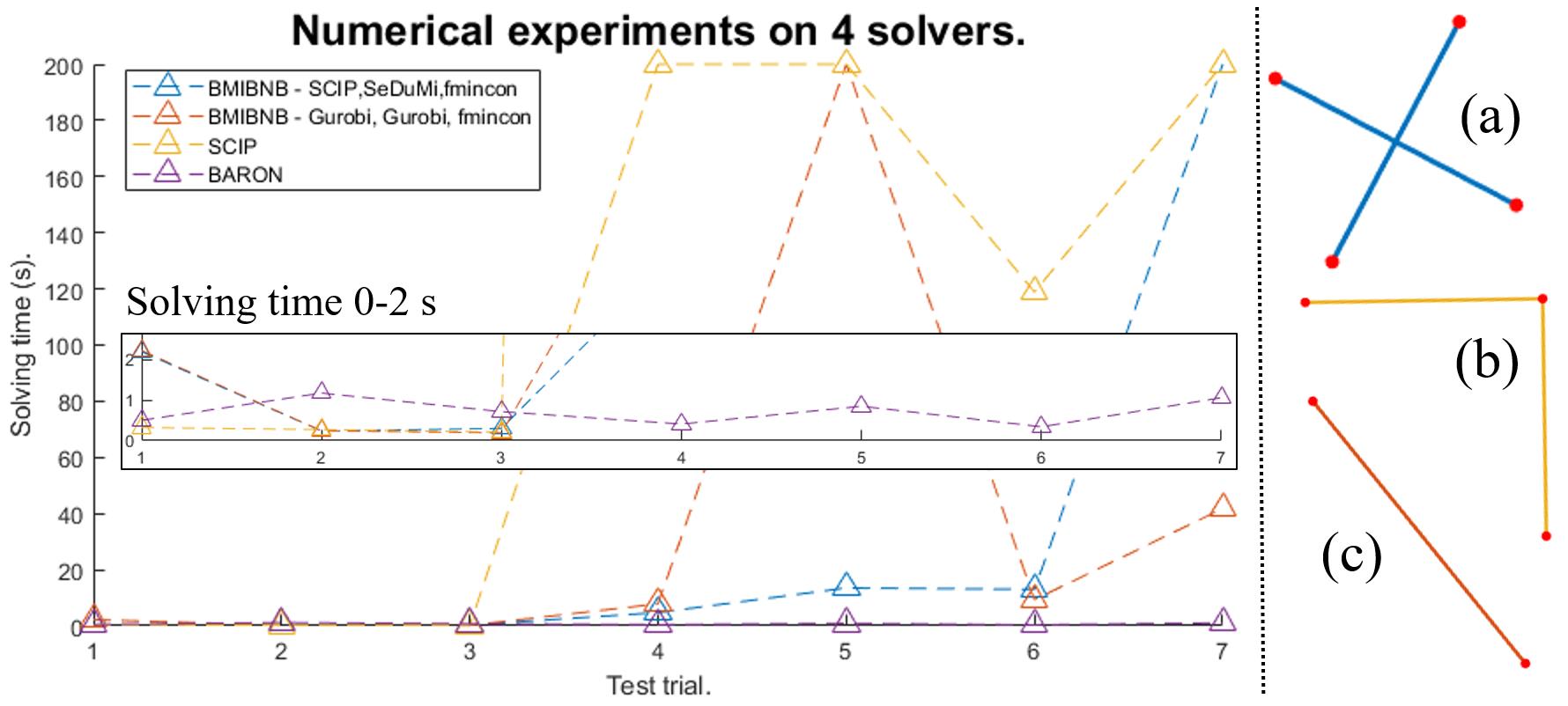}
  \caption{Left: Solver performance on preliminary test suite. Note that the optimization timeout has been set to 200 s. Right: Preliminary \textit{road assets} used in the benchmark.}\label{fig:toytest}
\end{figure}

We run numerical experiments on BMIBNB, SCIP, BARON to choose the one we will use to proceed. Since BMIBNB is a BNB technique that relies on external solvers, we test it with two solver configurations: a) LP: SCIP, lower bound: SeDuMi, upper bound: fmincon; b)  LP: Gurobi, lower bound: Gurobi, upper bound: fmincon. The preliminary testing suite is based on model \ref{model:phase1} with three trivial \textit{road asset} candidates. We initiate 7 tests with each one challenging the solver to map one non-empty subset of those assets into a given squared space. One single asset is selected in trial 1 to 3; two are involved in trial 4 to 6; and finally all three assets are selected in trial 7. Solving time for each tested solver is recorded and plotted (see Figure \ref{fig:toytest}).

We set the timeout to 200s since the tests are based on trivial \textit{road assets} with no more than four nodes each. Such timeout turns out to be more than enough because BARON passed every trial in around 1 second. Solving times reaching 200s indicate failed optimization due to timeout. Comparing the performance of two BMIBNB configurations on trial 5 and 7, we notice that the performance of BMIBNB highly depends on the utilized external solvers. SCIP ranks the last with 3 failed trials. We choose BARON to proceed with due to its obvious advantage over alternatives.

\subsection{Model Scalability and Complexity Analysis}

\begin{figure}[!ht]
  \centering
  \includegraphics[width=1\textwidth]{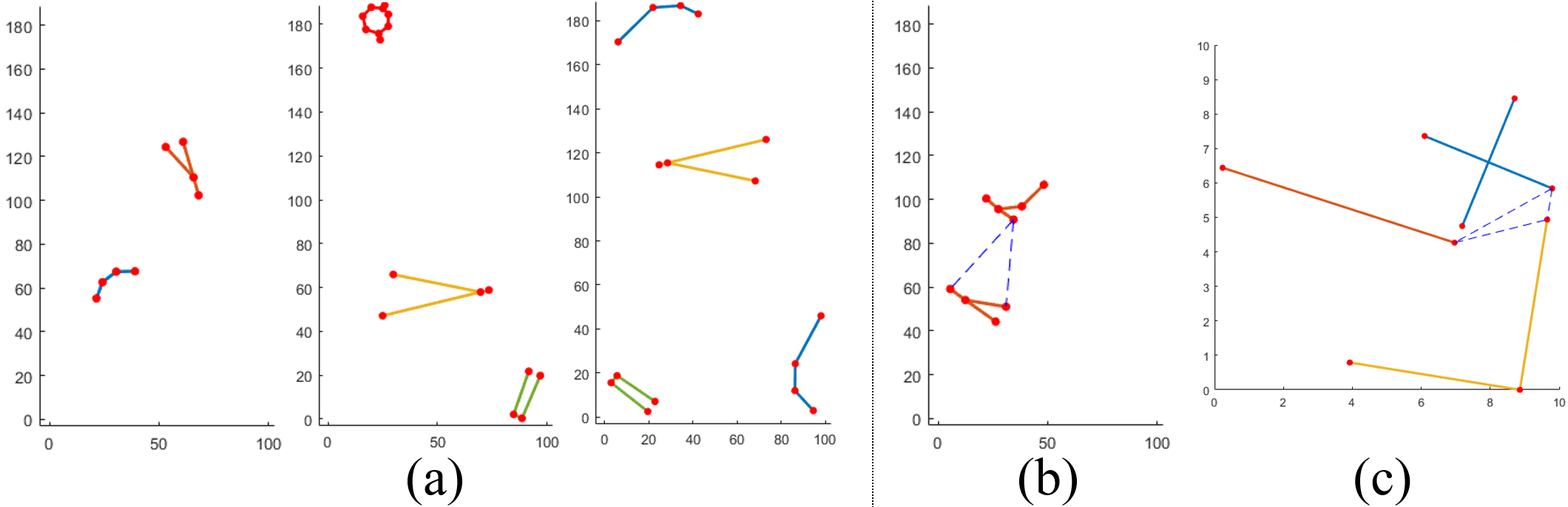}
  \caption{Examples of (a) feasible mapping search (phase 1), (b)-(c) flexibility optimization (phase 2). Blue dashed lines indicate the transition roads added by Xcity design process.}\label{fig:scale_test}
\end{figure}

Using BARON, we explore the scalability of our optimization models with \textit{road assets} $A_M$ which are extracted from Mcity road map. We gradually increase the number of available \textit{road assets} until BARON fails to solve within 100s. First, we run feasible mapping searches (phase 1) with 1, 2, and 3 randomly selected \textit{road assets}. BARON passed all tests with average solving time among 10 trials being 0.1678s, 2.9325s, and 30.9906s respectively for each \textit{road asset} quantity. Then, we run flexibility optimizations (phase 2) with all three preliminary \textit{road assets} and most valuable subsets $A^*$ from previous tests. See Figure \ref{fig:scale_test} for example outputs.

Having solved smaller problems successfully, BARON fails in searching for a feasible mapping for more than four assets in phase 1, or in optimizing transition flexibility of more than three assets in phase 2. In both cases, BARON fails to secure any feasible solution during preprocessing. The upper bound consequently stays at default value ($10^{51}$) after more than 5 hours' solving. Noticing the significant performance drop in solving, we analyze our model complexity and present the results in this section. We calculate the number of nonlinear constraints and binary variables introduced by each constraint set (see Table \ref{tab:complex}). Some notations are explained as follow:
\begin{itemize}
\item $n_i$: number of nodes in \textit{road asset} $a_i$.
\item $\ell$/$\ell_i$: all \textit{internal segments} / \textit{internal segments} in $a_i$
\item $N_\Delta$: cardinality of $\Delta$.
\item $X = \sum_{a_i,a_j}N_{\beta(a_i)}N_{\beta(a_j)}$: the number of all transition road candidates.
\end{itemize}

\begin{table}[htbp]
  \centering
  \caption{Model Complexity Analysis}
    \begin{tabular}{cllll}
    \toprule
    \multicolumn{1}{l}{Source} & Constraints / Binaries  & Quantity & Complexity & Type \\
    \midrule
    \multicolumn{1}{l}{\multirow{2}[2]{*}{SACS}} & Dist Cons. \ref{eq:d1}-\ref{eq:d3} & $\sum 2n_i-3$ & $O(N)$ & Quadratic\\
          & Orient Cons. (\ref{eq:ornt}) & $\sum n_i-2$ & $O(N)$ & Bilinear \\
    \midrule
    \multicolumn{1}{l}{\multirow{2}[2]{*}{CACS}} & Intersection Ineq. \ref{cons:ca}, \ref{model:phase1-2} & $2\left(C_{N_\ell}^2-\sum_{a_i}C_{N_{\ell_i}}^2\right)$ & $O(N_\ell^2)$ & Polynomial \\
          & Auxiliary Bin. \ref{cons:ca} & $2\left(C_{N_\ell}^2-\sum_{a_i}C_{N_{\ell_i}}^2\right)$ & $O(N_\ell^2)$ & Element-wise \\
    \midrule
    \multirow{2}[2]{*}{Connect.} & Intersection Ineq. \ref{model:phase2-4} & $2X(N_\ell-2)$ & $O(N_A^2N_\ell)$ & Polynomial \\
          & Auxiliary Bin. \ref{model:phase2-2}-\ref{model:phase2-4} & $C_{N_A}^2+X+2X(N_\ell-2)$ & $O(N_A^2N_\ell)$ & Element-wise \\
    \bottomrule
    \end{tabular}%
  \label{tab:complex}%
\end{table}%
Analytical result shows that \textit{road asset} quantity and \textit{internal segment} quantity are both major factors contributing to model complexity. The usage of excessive binary variables and nonlinear constraints requires a significant amount of branching process and relaxation solving, which has become the major computational burden. Notably, \textit{road asset} quantity, \textit{internal segment} quantity, and constraint type are all determined by problem construction and formulation, thereby fixed for a given problem. Therefore, we will work on problem decomposition to address the computational difficulty.

\section{Conclusion}


In this study, we proposed the first procedural scenario-based optimization approach for CAV proving ground design, and refer to it as an ``Xcity'' design problem. The notion of traffic encounters is employed to define \textit{road assets} as the fundamental construction elements of a CAV testbed. This infrastructure enables the use of scenario-based capability evaluation and direct-connectivity-based flexibility evaluation for CAV proving grounds. Then the Xcity design problem is formulated as general nonlinear optimization problems to fulfill the demanding fidelity requirement of CAV testing and validation tasks. We also presented a two-phase optimization model and demonstrated the design procedures using limited number of \textit{road assets}. Our numerical results showed the effectiveness of the formulation to guide Xcity constructions strategically. Due to the exponential computational complexity of the resulting formulation, the use of decomposition methods signifies our future direction to extend the tractability of the proposed formulation.

\section{Author Contribution Statement}
The authors confirm contribution to the paper as follows: study conception and design: Rui Chen, Mansur Arief, Ding Zhao; data collection: Rui Chen; analysis and interpretation of results: Rui Chen, Mansur Arief, Ding Zhao; draft manuscript preparation: Rui Chen, Mansur Arief, Ding Zhao. All authors reviewed the results and approved the final version of the manuscript.

\newpage

\printbibliography
\end{document}